\documentclass[]{article}

\usepackage[letterpaper]{geometry}
\usepackage{amta2020}
\usepackage{times}
\usepackage{url}
\usepackage{latexsym}
\usepackage{natbib}
\usepackage{layout}

\usepackage{amsmath}
\usepackage{lettrine}
\usepackage{latexsym}
\usepackage{url}
\usepackage{subcaption}
\usepackage{tikz} 
\usepackage{todonotes}
\usepackage{xcolor}
\usepackage{tabularx}
\usepackage[utf8]{inputenc}
\usepackage{booktabs} 
\usepackage{amssymb}
\usepackage{color}

\begin{document}

\amtaHeader{x}{x}{xxx-xxx}{2020}{Dynamic Masking for Online SLT}{Y. Yao and B. Haddow}
\title{\bf Dynamic Masking for Improved Stability in Online Spoken Language Translation}

\author{
        \name{\bf Yuekun Yao} \hfill \addr{ykyao.cs@gmail.com}\\
       \name{\bf Barry Haddow} \hfill \addr{bhaddow@staffmail.ed.ac.uk}\\
        \addr{School of Informatics, University of Edinburgh,  10 Crichton Street, Edinburgh, Scotland}
}

\maketitle
\pagestyle{empty}

\begin{abstract}
For spoken language translation (SLT) in live scenarios such as conferences, lectures and meetings,
it is desirable to show the translation to the user as quickly as possible, avoiding an annoying
lag between speaker and translated captions. In other words, we would like low-latency, online SLT.
 If we assume a pipeline of automatic speech recognition (ASR)
and machine translation (MT) then a simple but effective approach to online SLT is to pair an online ASR system, with a 
\emph{retranslation strategy}, where the MT system retranslates every update received from ASR. However
this can result in annoying ``flicker'' as the MT system updates its translation. 
A possible solution is to add a fixed delay, or ``mask'' to the
the output of the MT system, but a fixed global
mask re-introduces undesirable latency to the output. We introduce a method for dynamically
determining the mask length,
which provides a better latency--flicker trade-off curve than a fixed mask,  without affecting translation quality.
\end{abstract}

\section{Introduction}

A common approach to Spoken Language Translation (SLT) is to use a cascade (or pipeline) consisting of automatic speech 
recognition (ASR) and machine translation (MT). In a live translation setting, such as a lecture or conference, we would like
the transcriptions or translations to appear as quickly as possible, so that they do not ``lag'' noticeably behind the speaker.
 In other words, we wish to  minimise the latency of the system.
Many popular ASR toolkits can operate in an \emph{online} mode, where the transcription is produced
incrementally, instead of waiting for the speaker to finish their utterance.
However online MT is less well supported, and
  is complicated by the reordering which is often necessary in translation,
and by the use of encoder-decoder models which assume sight of the whole source sentence.

%

Some systems for online SLT rely on the \emph{streaming} approach to translation, perhaps inspired
by human interpreters. In this approach,  the MT system is
modified to translate incrementally, and on each update from ASR it will decide whether to update its
translation, or wait for further ASR output \citep{DBLP:journals/corr/ChoE16,ma-etal-2019-stacl,zheng-etal-2019-simpler,zheng-etal-2019-simultaneous,arivazhagan-etal-2019-monotonic}.

The difficulty with the streaming approach is that the system has to choose between committing to a particular
choice of translation output, or waiting for further updates from ASR, and does not have the option to revise
an incorrect choice. Furthermore, all the streaming approaches referenced above require specialised training
of the MT system, and modified inference algorithms. 

%
%
%
%
%
To address the issues above, we construct our online SLT system using the \textit{retranslation} approach
 \citep{niehues2018inter,arivazhagan2019retranslation}, which is less studied but more straightforward. It can be
implemented using any standard MT toolkit (such as Marian \citep{mariannmt}  which is highly optimised
for speed) and using the latest advances in text-to-text translation.
 The idea of retranslation  is that we  produce a new translation of the current sentence every time a
 partial sentence is received from the ASR system.
Thus, the translation of each sentence prefix is independent and can be directly handled by a standard MT system.
\citet{arivazhagan2020retranslation} directly compared the retranslation and streaming approaches and 
found the former to have a better latency--quality trade-off curve.

When using a completely unadapted MT system with the retranslation approach, however, there are at least two problems we need to consider:
\begin{enumerate}
  \item MT training data generally consists of full sentences, and systems may perform poorly on partial sentences
  \item When MT systems are asked to translate progressively longer segments of the conversation, they may introduce radical
  changes in the translation as the prefixes are extended. If these updates are displayed to the user, they will introduce
  an annoying ``flicker'' in the output, making it hard to read.
\end{enumerate}

We illustrate these points using the small example in Figure \ref{fig:de-example}. When the MT system receives the
first prefix (``Several'') it attempts to make a longer translation, due to its bias towards producing sentences.
When the prefix is extended (to ``Several years ago''), the MT system completely revises its original translation
hypothesis. This is caused by the differing word order between German and English.

\begin{figure}[htbp]
\centering
\begin{tabular}{lcl}
Several & $\longrightarrow$ & Mehrere Male \\
        &                   & \emph{Several times} \\
\\
Several years ago & $\longrightarrow$ & Vor einigen Jahre \\
        &                   & \emph{Several years ago} \\
\end{tabular}

\caption{Sample translation with standard en$\rightarrow$de MT system. We show the translation output, and its
back-translation into English.}
\label{fig:de-example}
\end{figure}

The first problem above could be addressed by simply adding sentence prefixes to the training data of the 
MT system. 
In our experiments we found that using prefixes  in training could improve translation
of partial sentences, but required careful mixing of data,
and even then performance of the model trained on truncated sentences was often worse on full sentences. 

A way to address both problems above is with an appropriate retranslation strategy.
In other words, when the
MT system receives a new prefix, it should decide whether to transmit its translation in full, partially, or 
wait for further input, and the system can take into account translations it previously produced.
A good retranslation strategy will address the second problem above (too much flickering as translations
are revised) and in so doing so address the first (over-eagerness to produce full sentences).

In this paper, we focus on the retranslation methods
 introduced by \citet{arivazhagan2019retranslation} -- mask-$k$ and 
biased beam search. The former is a delayed output strategy which does not affect overall quality, reduces flicker,
 but
can significantly increase the latency of the  translation system. The latter alters the beam search
to take into account the translation of the previous prefix, and  is  used to reduce flicker without
influencing latency much, 
but can also damage translation quality. 

Our contribution in this paper is to show that by using a straightforward method to predict how much to 
mask (i.e. the
value of $k$ in the mask-$k$ strategy) we obtain a more optimal trade-off of
 flicker and latency than is possible with a fixed mask. We show that for many prefixes the mask can be 
 safely reduced.
 We achieve this by having the system make  probes of possible extensions to the source prefix, and observing how 
 stable the translation of these probes is -- instability in the translation requires a larger mask.
 Our method requires no  modifications to the underlying MT system, and has no effect on translation quality.

\section{Related Work}
\label{sec:lfq}

Early work on incremental MT used prosody \citep{bangalore-etal-2012-real} or lexical cues \citep{rangarajan-sridhar-etal-2013-segmentation} to make the translate-or-wait decision.
The first work on incremental neural MT used confidence to decide whether to 
wait or translate \citep{DBLP:journals/corr/ChoE16}, whilst in \citep{gu-etal-2017-learning} they learn the
translation schedule with reinforcement learning. 
In \citet{ma-etal-2019-stacl}, they address simultaneous translation using a  
transformer \citep{vaswani2017attention} model with a modified attention
mechanism,  which is trained on prefixes. They introduce the idea of wait-$k$, where the
translation does not consider the final $k$ words of the input. This work was 
extended by \citet{zheng-etal-2019-simultaneous,zheng-etal-2019-simpler},
where a ``delay'' token is added to the target
vocabulary so the model can learn when to wait, through being trained by imitation learning.
The MILk attention \citep{arivazhagan-etal-2019-monotonic} also provides a way of learning
the translation schedule along with the MT model, and is able to directly optimise
the latency metric.

In contrast with these recent approaches, retranslation strategies \citep{niehues2018inter} allow the use of a 
standard MT toolkit, with little modification, and so are able to leverage all 
the performance and quality optimisations in that toolkit. \citet{arivazhagan2019retranslation} pioneered
the retranslation system by combining a strong MT system with two simple yet effective strategies: biased beam search and mask-$k$. Their experiments show that the system can achieve low flicker and latency without losing much performance. 
In an even more recent paper, \citet{arivazhagan2020retranslation} further combine their retranslation system with prefix training and make comparison with current best streaming models (e.g. MILk and wait-$k$ models), showing that such a retranslation system is a strong option for online SLT.


\section{Retranslation Strategies}

Before introducing our approach, we describe the two retranslation strategies
introduced in \cite{arivazhagan2019retranslation}: mask-$k$
and biased beam search.

 The idea of mask-$k$ is simply that the MT system does not transmit
 the last $k$ tokens of its output -- in other
words it masks them. Once the system receives a full sentence, it transmits the translation in full, 
without masking. The value of $k$ is set globally and can be tuned to reduce the amount of flicker, at the
cost of increasing latency. \cite{arivazhagan2019retranslation}  showed good results for a mask of 
10, but of course for short sentences a system with such a large mask  would not produce
any output until the end of the sentence.

In biased beam search, 
a small modification is made  to the translation algorithm, changing the search objective.
The technique aims to reduce flicker by ensuring that the translation produced by the MT system
stays closer to the translation of the previous (shorter) prefix. Suppose that $S$ is a source prefix, $S'$ is
the extension of that source prefix provided by the ASR,
 and $T$ is the translation of $S$ produced by the system (after masking).
 Then to create the translation $T'$ of $S'$, biased beam search substitutes the model
probability $p(t'_i|t'_{<i}, S')$ with the following expression:
\begin{equation}
p^B(t'_i|t'_{<i}, S') = (1 - \beta) \cdot  p(t'_i|t'_{<i}, S') + \beta \cdot \delta (t'_i, t_i)  
\label{eqn:bias}
\end{equation}
where $t'_i$ is the $i^{\text{th}}$ token of the translation hypothesis $T'$, and $\beta$ is a weighting 
which we set to 0 when  $t_{<i} \neq t'_{<i}$. In other words, we  interpolate the translation model
with a function that keeps it close to the previous translation, but stop applying the biasing once the new
translation diverges from the previous one. 

As we noted earlier, biased beam search can degrade the quality of the translation, and we show experiments
to illustrate this in Section~\ref{sec:expts-biased}. We also note that biased beam search assumes that the
ASR simply extends its output each time it updates, when in fact ASR systems may rewrite their output.
Furthermore, biased beam search requires modification of the underlying inference algorithm
(which in the case of Marian is written in highly optimised, hand-crafted GPU code), removing one
of the advantages of the retranslation approach (that it can be easily applied to standard MT systems).

\section{Dynamic Masking}
\label{sec:dynamic}
In this section we introduce our improvement to the mask-$k$ approach, which uses a variable mask, that is 
set at runtime. The problem with using a fixed mask, is that there are many time-steps where the system
is unlikely to introduce radical changes to the translation as more source is revealed, and on these occasions
we would like to use a small mask to reduce latency. However the one-size-fits-all mask-$k$ strategy
does not allow this variability. 

The main idea of dynamic masking  is to \emph{predict} what the next source word will be, and check what effect this would have on the translation. If this
changes the translation, then we mask, if not we output the full translation.

\begin{figure*}[ht!]
\begin{center}
\begin{tikzpicture}
\tikzstyle{box} = [rectangle, minimum height=0.7cm, minimum width=1.5cm, draw, rounded corners, anchor=center];
\tikzstyle{source} = [box];
\tikzstyle{target} = [box];
\tikzstyle{arr} = [thick,->,>=stealth, shorten >=1mm,shorten <=0mm,];

\node (prefix) [source] {$S = a~b$};
\node (asr) [left = 1.5cm of prefix] {ASR};
\node (eprefix) [below = 2cm of prefix.center, source ] (eprefix) {$S' = a~b~c$};

\node (tprefix) [right = 4cm of prefix.center, target]  {$T = p~q~r$};
\node (etprefix) [below = 2cm of tprefix.center, target]  {$T' = p~q~s~t$};

\node (lcp) [below right = 1cm and 2cm of tprefix.center, circle, draw, minimum width=0.1cm] {};
\node (final) [right = 2cm of lcp, target] {$T^* = p~q$};

\draw[arr] (asr) -- (prefix);

\draw[arr] (prefix) -- (eprefix) node[midway, left] {extend};
\draw[arr] (prefix) -- (tprefix) node[midway, above] {translate};
\draw[arr] (eprefix) -- (etprefix) node[midway, below] {translate};

\draw[arr] (tprefix) -- (lcp);
\draw[arr] (etprefix) -- (lcp);

\draw[arr] (lcp) -- (final) node[midway, above] {LCP};

\end{tikzpicture}
\end{center}
\caption{The source prediction process. The string $a~b$ is provided by the ASR system. The MT system then
produces translations of the string and its extension, compares them, and outputs the longest common prefix 
(LCP) }
\label{fig:source-pred}
\end{figure*}
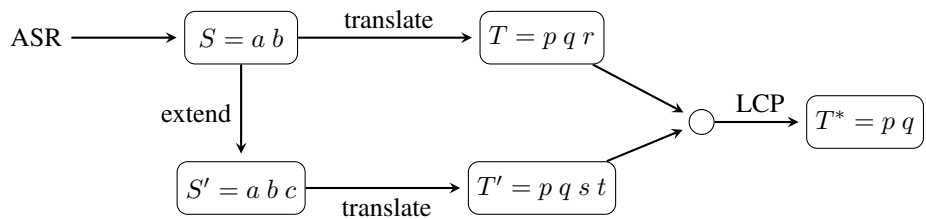

More formally, we suppose that we have a 
source prefix $S = s_1\ldots s_p$, a source-to-target translation system, and a function $pred_k$,
 which can predict 
the next $k$ tokens following $S$. We translate $S$ using the 
translation system to give a translation hypothesis $T = t_1\ldots t_q$. We then use $pred_k$ to predict the tokens
following $s_p$ in the source sentence to give an extended source prefix $S' = s_1\ldots s_ps_{p+1}\ldots s_{p+k}$, and
translate this to give another translation hypothesis $T'$.
Comparing $T$ and $T'$, we select the longest common prefix $T^*$, and output this as the translation, 
thus masking the final $|T| - |T^*|$ tokens of the translation. If $S$ is a complete sentence, then we do not mask any of the output,
as in the mask-$k$ strategy.
The overall procedure is illustrated in Figure~\ref{fig:source-pred}.

 In fact, after initial experiments, we found it was more effective to refine our strategy, and not
mask at all if the translation after dynamic mask is a prefix of the previous translation. In this case we directly output the last translation.
In other words, we do not mask if $is\_prefix(T_{i}^{*}, T_{i-1}^{*})$ but instead
output  $T_{i-1}^{*}$ again, where $T_{i}^{*}$ denotes the masked translation for the $i$th ASR input. We also notice that this refinement does not give any benefit to the mask-$k$ strategy in our experiments. The reason 
that this refinement is effective is that the translation of the extended prefix can sometimes exhibit 
instabilities early in the sentence (which then disappear in a subsequent prefix). Applying a large mask 
in response to such instabilities increases latency, so we effectively ``freeze'' the output of the 
translation system until the instability is resolved. An example to illustrate this refinement can be found in 
Figure~\ref{tab:ex2}.

To predict the source extensions (i.e.\ to define the $pred_k$ function above), we first tried using a  language
model trained on the source text. This worked well, but we decided to add two simple strategies in order
to see how important it was to have good prediction. The extension strategies we include are:
\begin{description}
  \item[lm-sample] We sample the next token from a language model (LSTM) trained on the source-side of the parallel training data. We can choose $n$ possible extensions by choosing $n$ distinct samples.
  \item[lm-greedy] This also uses an LM, but chooses the most probable token at each step.
  \item[unknown] We extend the source sentence using the \textsc{unk} token from the vocabulary.
  \item[random] We extend by sampling randomly from the vocabulary, under a uniform distribution. As
  with lm-sample, we can generalise this strategy by choosing $n$ different samples.
\end{description}

\section{Evaluation of Retranslation Strategies}
\label{sec:eval}
Different approaches have been proposed to evaluate online SLT, so we explain and justify our approach here.
We follow previous work on retranslation \citep{niehues2018inter,arivazhagan2019retranslation,arivazhagan2020retranslation} and consider that the performance
of online SLT should be assessed according to three different aspects -- quality, latency and flicker.
All of these aspects are important to users of online SLT, but improving on one can have an adverse effect
on other aspects.
For example outputting translations as early as possible will reduce latency, but if these early outputs are incorrect then either they can be corrected (increasing flicker) 
or retained as part of the later  translation (reducing quality).
In this section we will define precisely how we measure these system aspects.  We assume that 
selecting the optimal trade-off between quality, latency and flicker is a question for the system deployer, 
 that can only be settled by user testing. 

\paragraph{Latency}
The latency of the MT system should provide a measure of the time between the MT system receiving 
input from the ASR, and it producing output that can be potentially be sent to the user.
A standard (text-to-text)  MT
system would have to wait until it has received a full sentence before it produces any output, which exhibits high latency. 
%
%

We follow  \citet{ma-etal-2019-stacl} by using a latency metric called average lag (AL), which measures the degree to which the output 
lags behind the input. This is done by averaging the difference between the number of words the system has output, and the number of words expected, given the length of the source prefix received, and the ratio between source and target length. 
Formally, AL for source and target sentences $S$ and $T$ is defined as:
$$
  \text{AL}(S, T) = \frac{1}{\tau} \sum_{t=1}^{\tau} g(t) - \frac{(t-1)|S|}{|T|}
$$
where $\tau$ is the number of target words generated by the time the whole source sentence
is received, $g(t)$ is the number of source words processed when the target hypothesis first reaches
a length of $t$ tokens.
In our implementation, we calculate the AL at token (not subword) level with
the
standard
tokenizer in sacreBLEU \citep{post-2018-call}, meaning that for Chinese output we calculate AL on characters.

This metric differs from the one used in \citet{arivazhagan2019retranslation}\footnote{In the presentation of
this paper at ICASSP, the authors used a  latency metric similar to the one used here, and different to the one
they used in the paper},
where latency is defined  as the mean time between a source word being received and the translation of that source 
word being finalised. We argue against this definition, because it conflates latency and flicker, since outputting
a translation and then updating is penalised for both aspects. The update is  penalised for flicker since
the translation is updated (see below) and it is penalised for latency, since the timestamp
of the initial output is ignored in the latency calculation.

\paragraph{Flicker}
The idea of flicker is to obtain a measure of the potentially distracting changes that are made to the MT output, as its ASR-supplied source sentence is extended. We assume that straightforward extensions of 
the MT output are fine, but changes which require re-writing of part of the MT output should result in a higher (i.e. worse) flicker score. Following \citet{arivazhagan2019retranslation}, we measure flicker using the normalised erasure (NE), which is defined as the minimum number of tokens that
must be erased from each translation hypothesis when outputting the subsequent hypothesis, normalised 
across the sentence. As with AL, we also calculate the NE at token level for German, and at character level
for Chinese.

\paragraph{Quality}
As in previous work \cite{arivazhagan2019retranslation}, quality is assessed by comparing the full sentence
output of the system against a reference, using a sentence similarity measure such as \textsc{bleu}.
We do not evaluate quality on prefixes, mainly because of the need for a heuristic to determine partial references. Further, quality evaluation on prefixes will conflate with evaluation of latency and flicker and thus we simply assume that if the partial 
sentences are of poor quality, that this will be reflected in the other two metrics (flicker and latency).
Note that our proposed dynamic mask strategy is only concerned with improving the flicker--latency
trade-off curve and has no effect on full-sentence quality, so MT quality measurement is not the focus of this paper.
Where we do require a measure of quality (in the assessment of biased beam search, which does
change the full-sentence translation) we use \textsc{bleu} as implemented by sacreBLEU \citep{post-2018-call}.


\section{Experiments} 
\label{sec:expts}

\subsection{Biased Beam Search and Mask-$k$}
\label{sec:expts-biased}

We first assess the effectiveness of biased beam search and mask-$k$ (with a fixed $k$), providing a more
complete experimental picture than in \citet{arivazhagan2019retranslation}, and demonstrating 
the adverse effect of biased beam search on quality. For these experiments we use
data released for the IWSLT MT task \citep{Cettolo2017OverviewOT}, in both English$\rightarrow$German
and English$\rightarrow$Chinese. We consider a simulated ASR system, which supplies the gold transcripts
to the MT
system one token at a time\footnote{A real online ASR system typically increments its hypothesis by adding
a variable number of tokens in each increment, and may revise its hypothesis. Also, ASR does not normally 
supply sentence boundaries, or punctuation, and these must be added by an intermediate component. Sentence 
boundaries may change as the ASR hypothesis changes. In this work we make simplifying assumptions about 
the nature of the ASR, in order to focus on retranslation strategies, leaving the question of dealing with
real online ASR to future work.}.

For training we use the TED talk data, with dev2010 as heldout and tst2010
as test set. The raw data set sizes are 206112 sentences (en-de) and 231266 sentences (en-zh).
 We preprocess using the Moses \citep{koehn-EtAl:2007:PosterDemo} tokenizer and truecaser
 (for English and German) and 
jieba\footnote{\url{https://github.com/fxsjy/jieba}} for Chinese. We apply
BPE \citep{sennrich-haddow-birch:2016:P16-12} jointly with 90k merge operations.
For our MT system, we use the transformer-base architecture \citep{vaswani2017attention} as implemented
by Nematus \citep{sennrich-etal-2017-nematus}. We use 256 sentence mini-batches, and a 4000 iteration warm-up
in training.

As we mentioned in the introduction, we did experiment with prefix training (using both alignment-based
and length-based truncation) and found that this improved the translation of prefixes, but generally
degraded translation for full sentences. Since prefix translation can also be improved using the 
masking and biasing techniques, and the former does not degrade full sentence translation, we only 
include experimental results when training on full sentences.

\begin{figure*}[ht!]
    \begin{subfigure}{0.5\textwidth}
       \includegraphics[width=\textwidth]{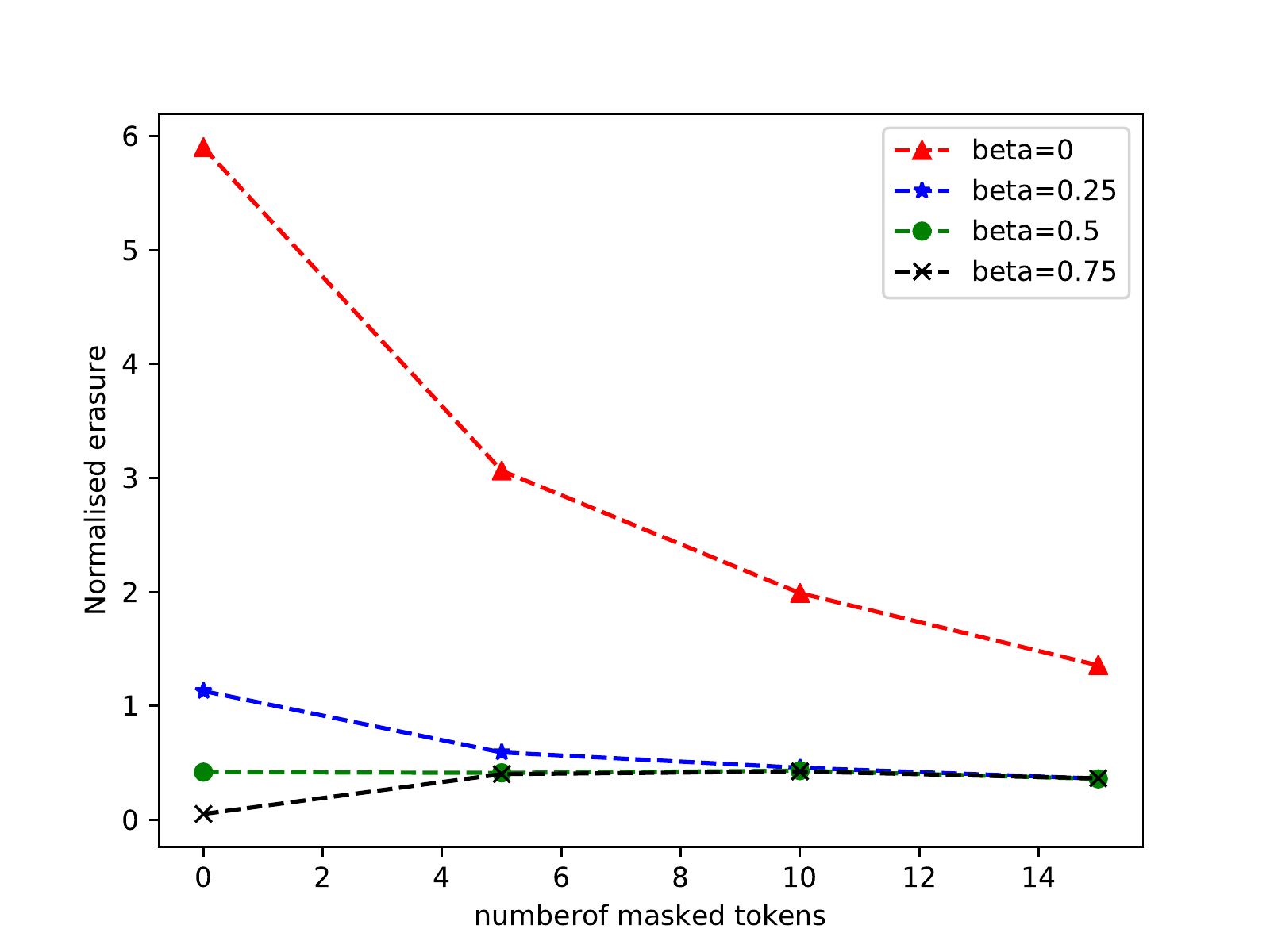}
        \caption{Masking versus flicker (measued by erasure)}
    \end{subfigure}
    \begin{subfigure}{0.5\textwidth}
        \includegraphics[width=\textwidth]{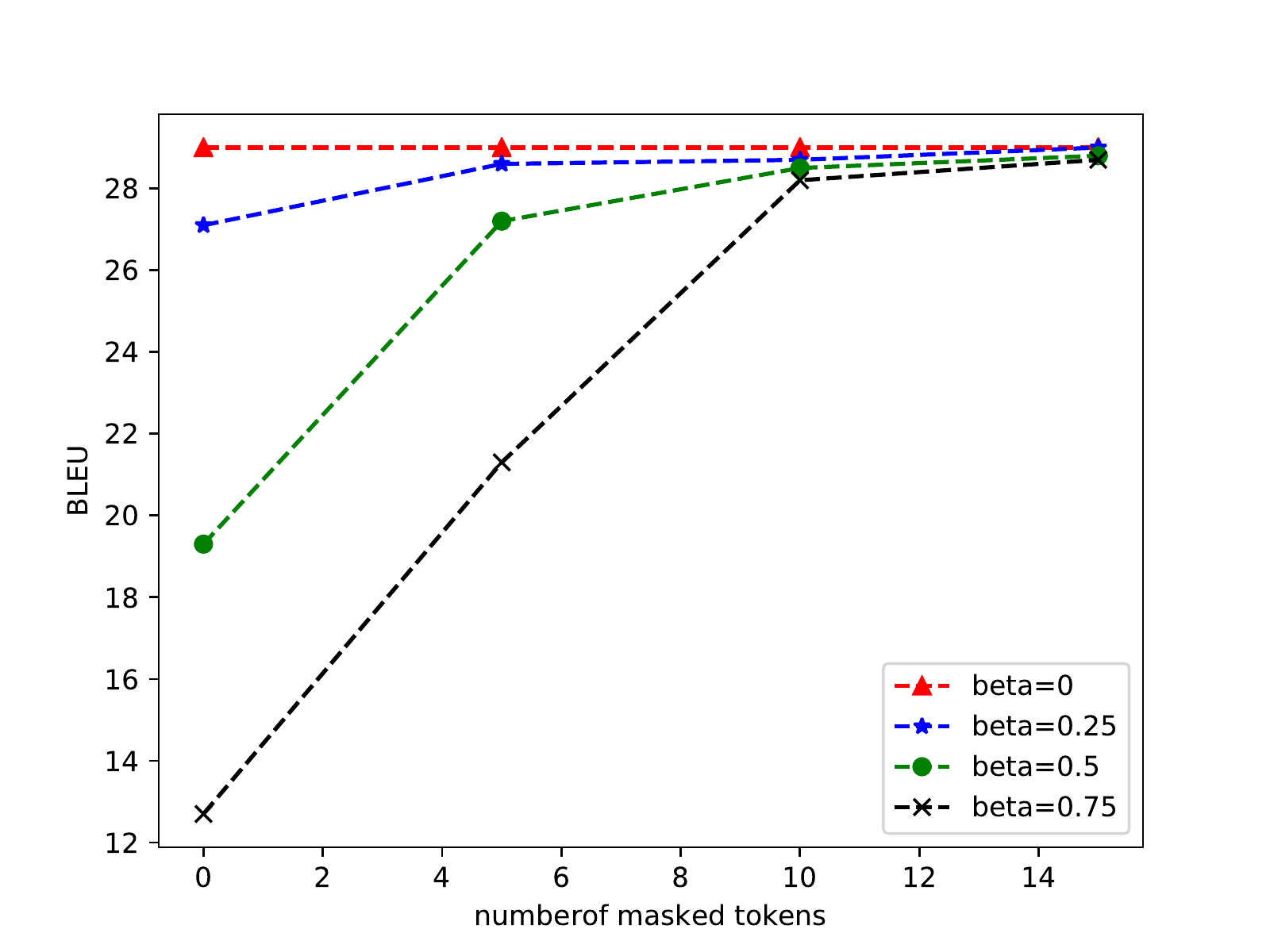}
        \caption{Masking versus quality (measured by \textsc{bleu}).}
    \end{subfigure}
     \begin{subfigure}{1.0\textwidth}
       \centering
       \includegraphics[width=0.5\textwidth]{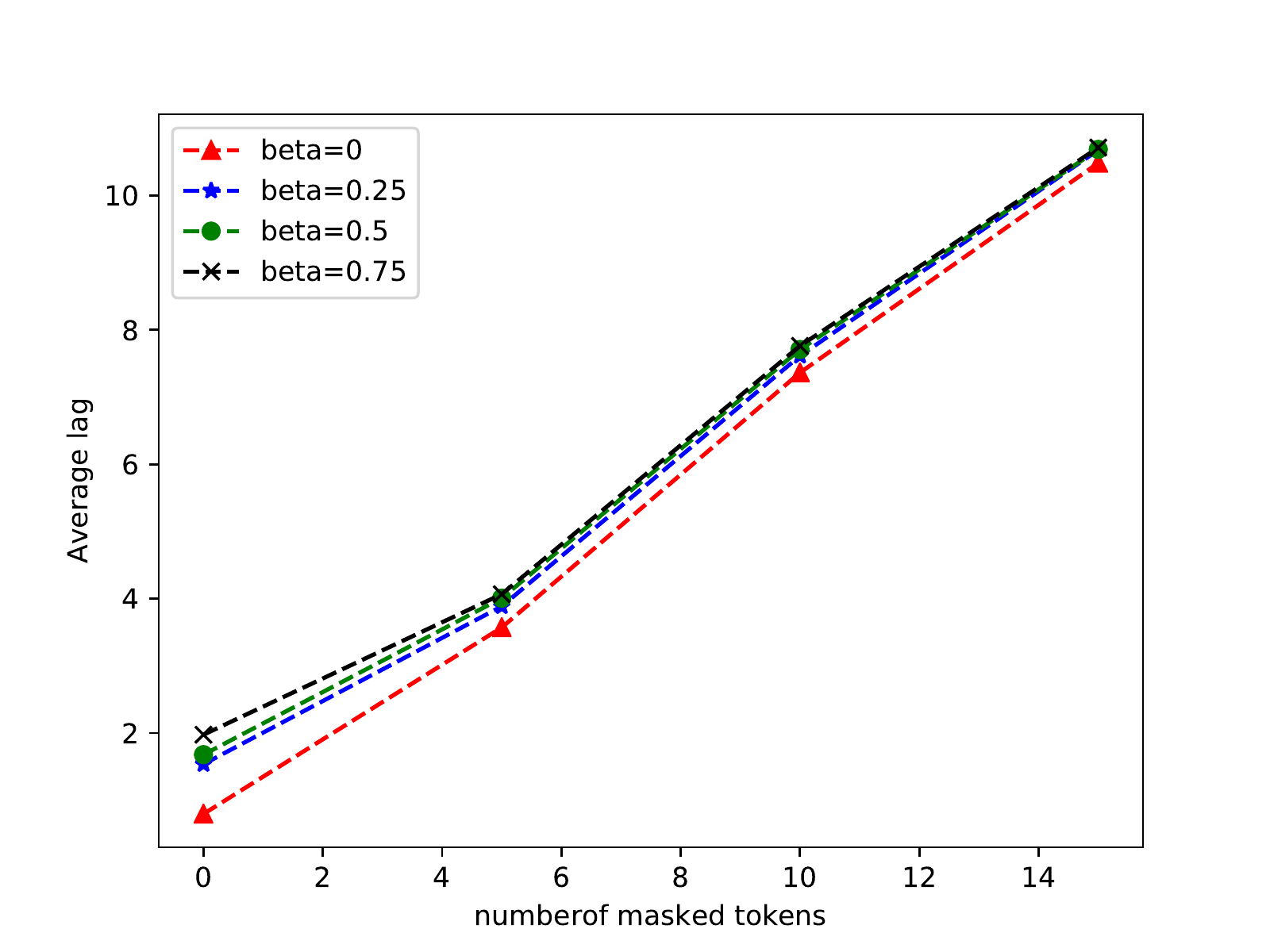}
        \caption{Masking versus  latency (measured by average lagging).}
    \end{subfigure}
    \caption{Effect of varying mask-$k$, at different values of the biased beam search interpolation
    parameter, on the three measures proposed in Section \ref{sec:eval}.}
    \label{fig:deen}
\end{figure*}

In Figure \ref{fig:deen} we show the effect of varying $\beta$ and $k$ on our three evaluation measures,
for English$\rightarrow$German.  

Looking at  Figure \ref{fig:deen}(a) we notice that biased beam search has a strong impact in reducing flicker
 (erasure)
at all values of $\beta$. However the problem with this approach is clear in Figure \ref{fig:deen}(b), 
where we can see the reduction in \textsc{bleu} caused by this biasing. This can be offset by increasing
masking, also noted by \citet{arivazhagan2019retranslation}, but as we show in Figure \ref{fig:deen}(c) this comes
at the cost of an increase in latency.

\begin{figure*}[ht!]
    \begin{subfigure}{0.5\textwidth}
       \includegraphics[width=\linewidth]{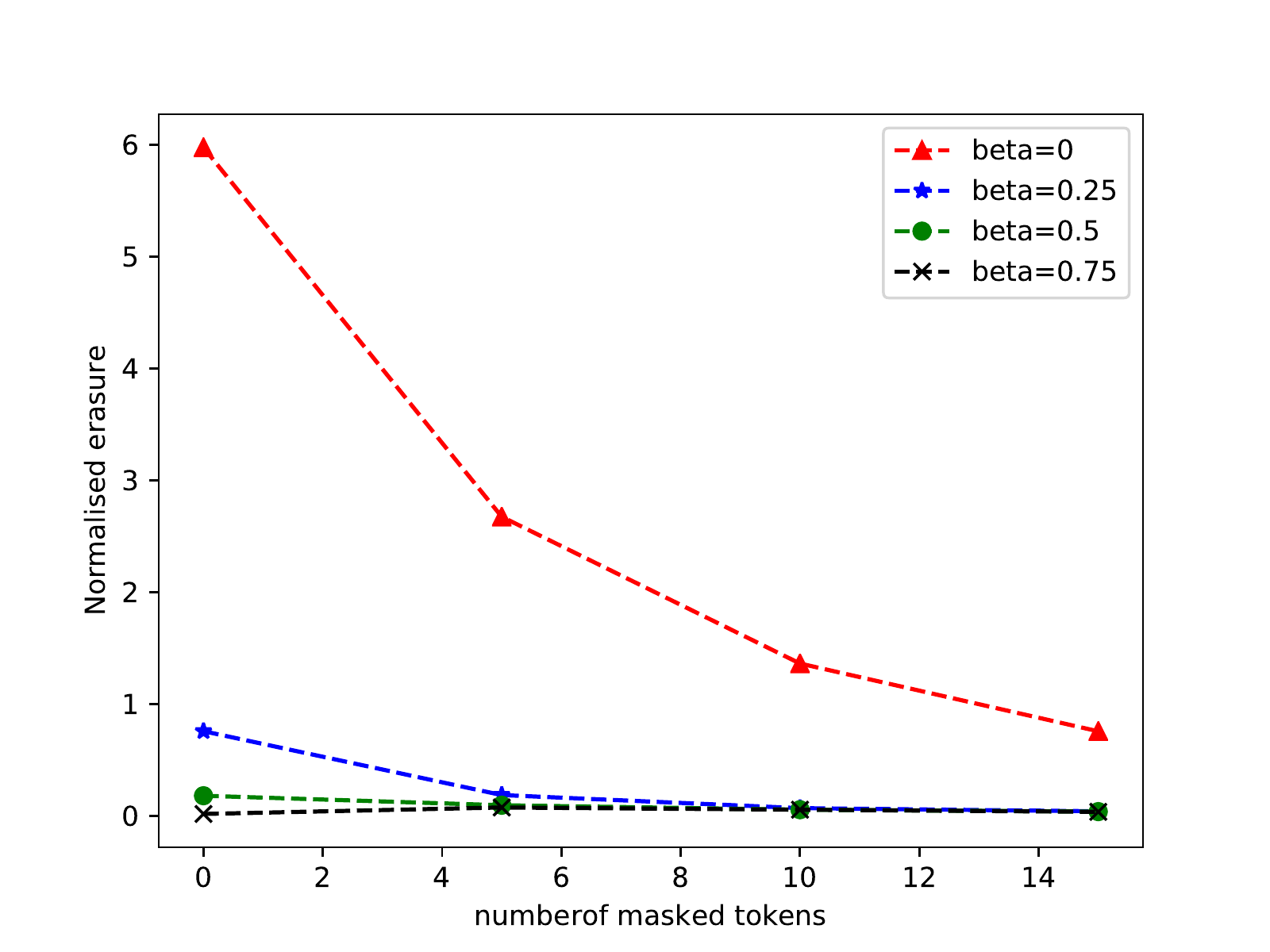}
        \caption{Masking versus flicker (measured by erasure)}
    \end{subfigure}
    \begin{subfigure}{0.5\textwidth}
        \includegraphics[width=\linewidth]{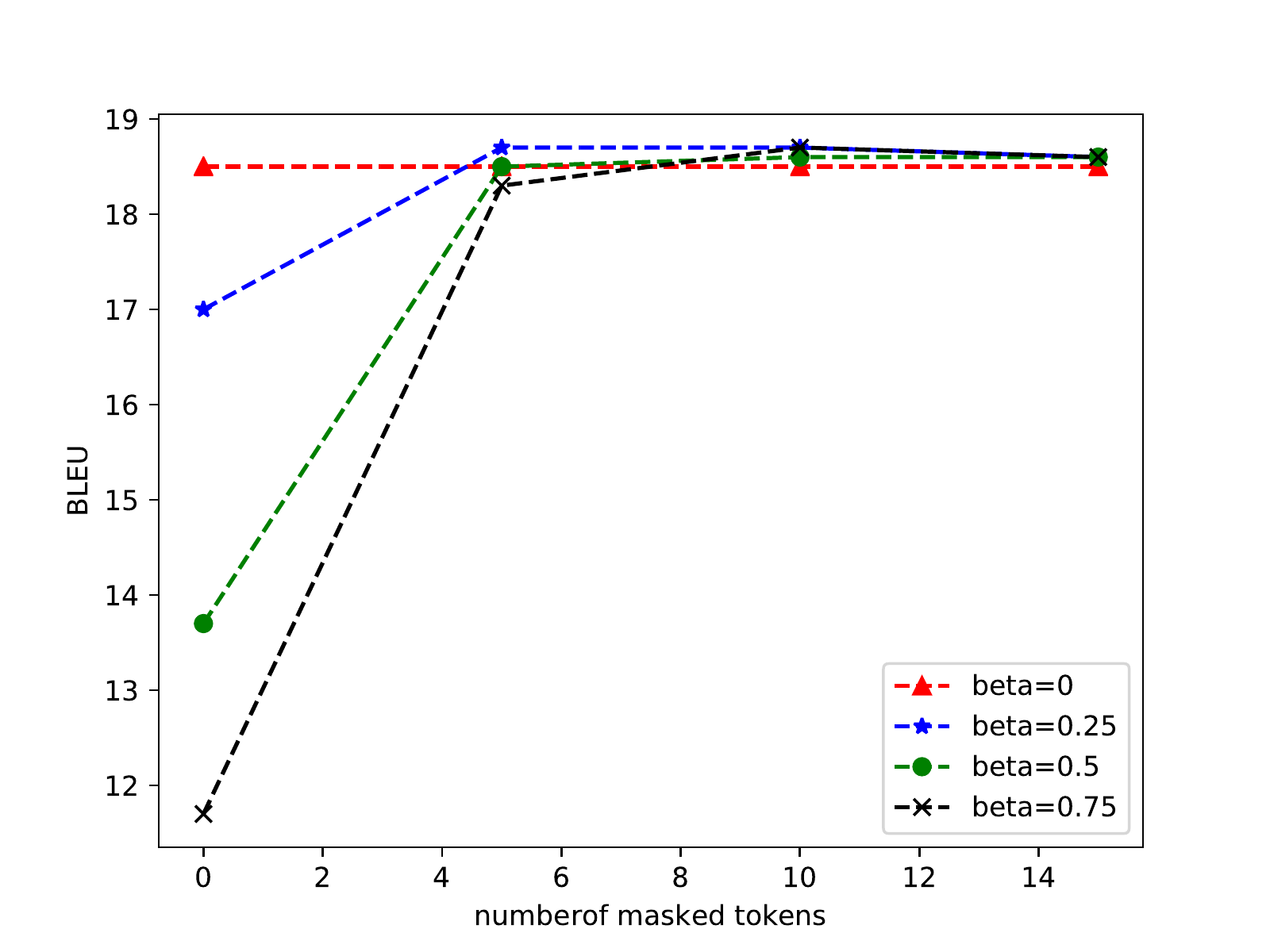}
        \caption{Masking versus quality (measured by \textsc{bleu}).}
    \end{subfigure}
     \begin{subfigure}{1.0\textwidth}
       \centering
       \centering
       \includegraphics[width=0.5\linewidth]{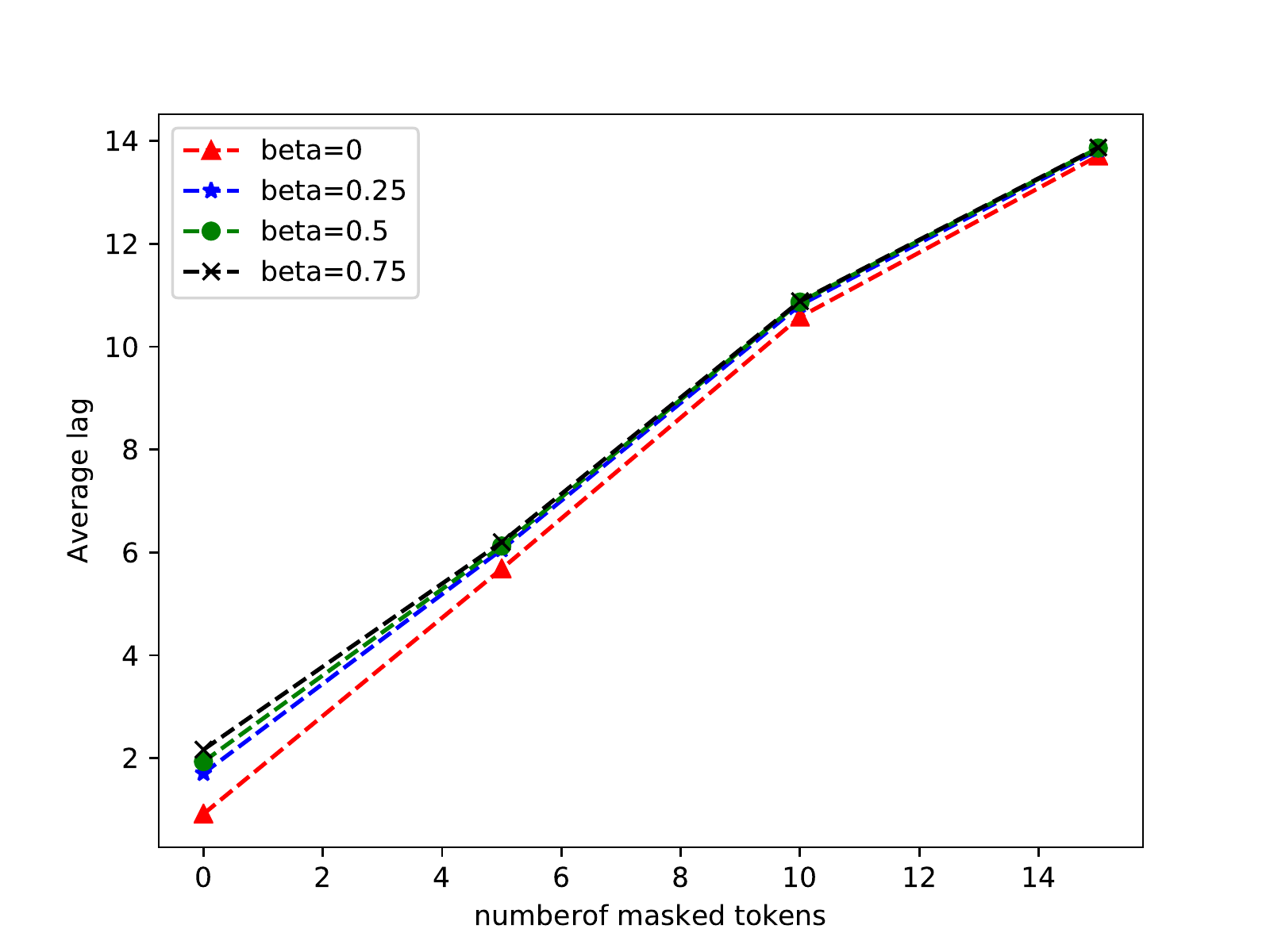}
        \caption{Masking versus  latency (measured by average lagging).}
    \end{subfigure}
    \caption{Effect of varying mask-$k$, at different values of the biased beam search interpolation
    parameter, on English-to-Chinese corpus}
    \label{fig:en-zh-mask-bias}
\end{figure*}

Our experiments with en$\rightarrow$zh show a roughly similar pattern, as shown in Figure \ref{fig:en-zh-mask-bias}. We find that
lower levels of masking are required to  reduce the detrimental effect on \textsc{bleu} of the biasing, but
latency increases more rapidly with masking.

\subsection{Dynamic Masking} 
\label{sec:expt-dynamic}

We now turn our attention to the dynamic masking technique introduced in Section \ref{sec:dynamic}. 
We use the same data sets and MT systems as in the previous section. To train the LM, we use the source
side of the parallel training data, and train an LSTM-based LM.

\subsubsection{Comparison with Fixed Mask}

\begin{figure*}[ht!]
    \begin{subfigure}{0.5\textwidth}
       \includegraphics[width=\linewidth]{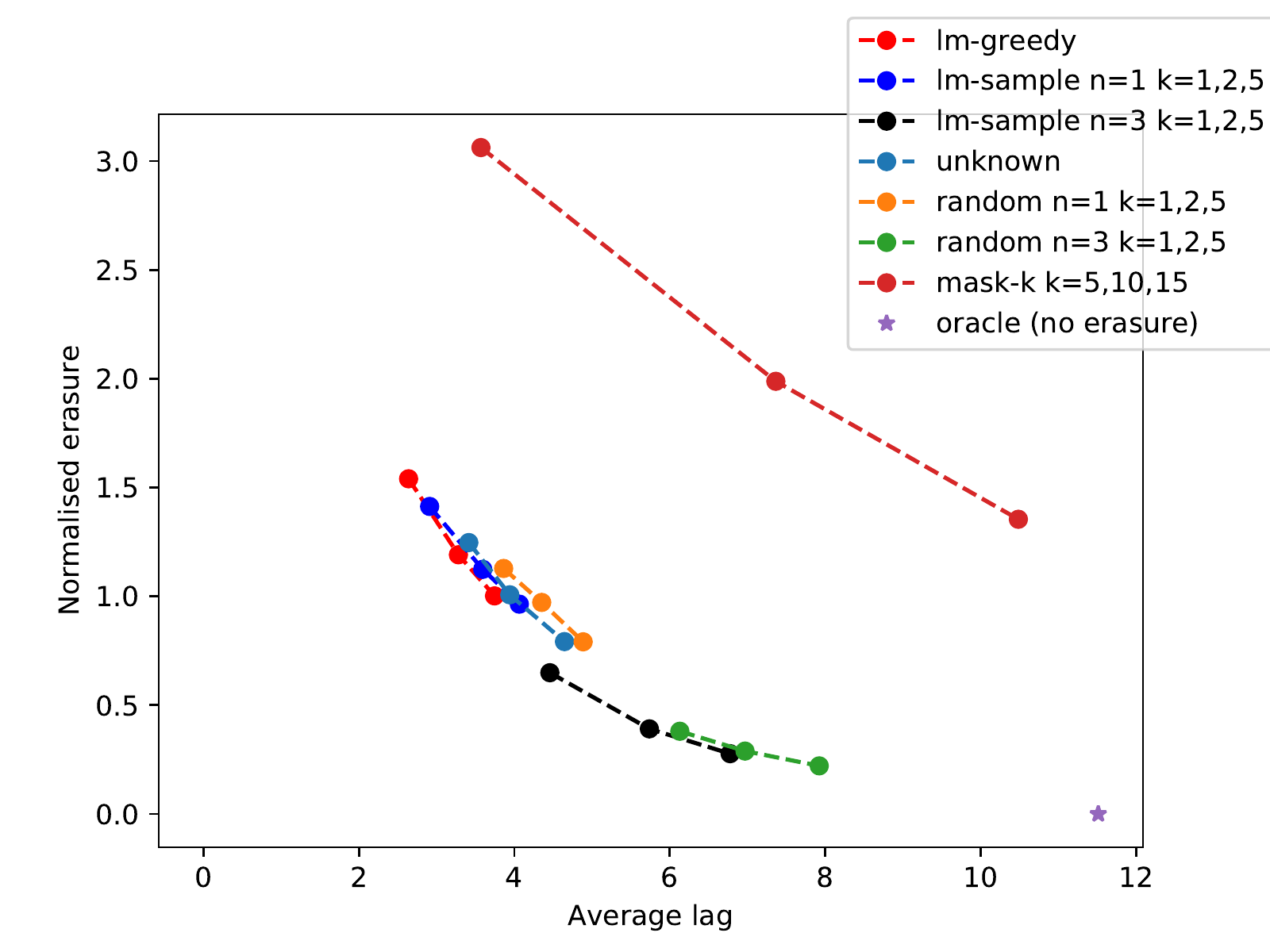}
        \caption{Dynamic Masking for en$\rightarrow$de MT system trained on full sentences (measured by AL and NE)}
    \end{subfigure}
    \begin{subfigure}{0.5\textwidth}
        \includegraphics[width=\linewidth]{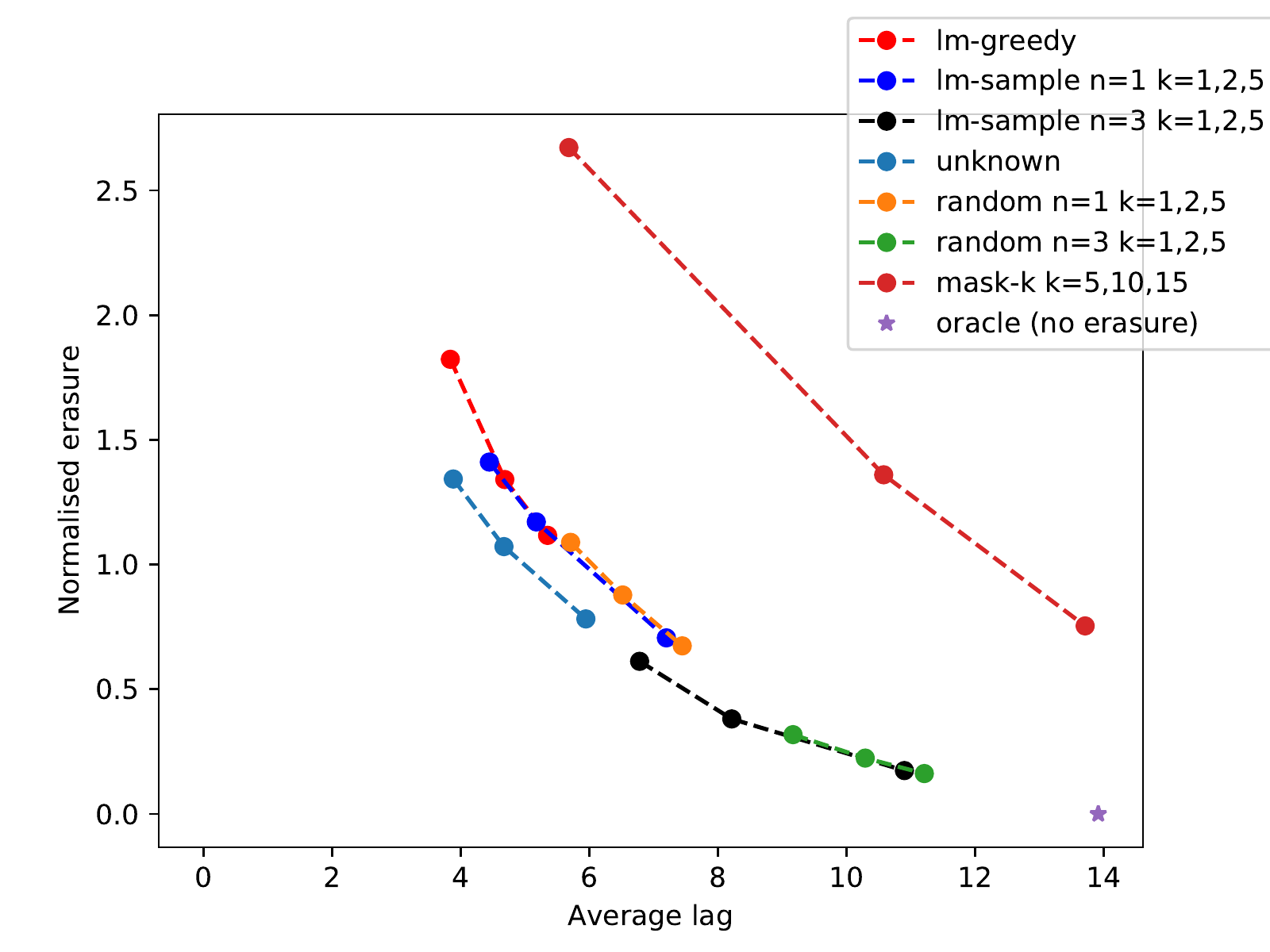}
        \caption{Dynamic Masking for en$\rightarrow$zh MT system trained on full sentences (measured by AL and NE)}
    \end{subfigure}
    \caption{Effect of dynamic mask with different source prediction strategies. The strategies are explained in Section 3, and the oracle means that we use the full-sentence translation to set the mask.}
    \label{fig:dyn_mask}
\end{figure*}

To assess the performance of dynamic masking, we measure latency and flicker as we vary the length 
of the source extension ($k$) and the number of source extensions ($n$).
We consider the 4 different extension strategies described
at the end of Section \ref{sec:dynamic}. We do not show translation quality since the translation of the 
complete sentence is unaffected by the
dynamic masking strategy.  The results for both en$\rightarrow$de and en$\rightarrow$zh
are shown in Figure \ref{fig:dyn_mask}, where we compare to the strategy of using 
a fixed mask-$k$. The oracle data point is where we use the full-sentence translations to set the 
mask so as to completely avoid flicker.

We observe from Figure \ref{fig:dyn_mask} that our dynamic mask mechanism improves over the fixed
mask in all cases, Pareto-dominating it on latency and flicker. Varying the source prediction strategy and
parameters appears to preserve the same inverse relation between latency and flicker, although
offering a different trade-off. 
 Using several random source predictions 
(the green curve in both plots) offers the lowest flicker, at the expense of high latency,
whilst the shorter LM-based predictions show the lowest latency, but higher flicker. 
We hypothesise that this  happens because extending the source
randomly along several directions, then comparing the translations is most likely to 
expose any instabilities in the translation and so lead
to a more conservative masking strategy (i.e.\ longer masks). In contrast, using a single LM-based predictions gives more plausible
extensions to the source, making the MT system more likely to agree on its translations of the prefix and its
extension, and so less likely to mask. Using several LM-based samples (the black curve) adds diversity and gives similar
results to random (except that random is quicker to execute).
The pattern across the two language pairs
is similar, although we observe a more dispersed picture for the en$\rightarrow$zh results. 

\begin{figure}
\begin{subfigure}{0.5\textwidth}
  \includegraphics[width=\linewidth]{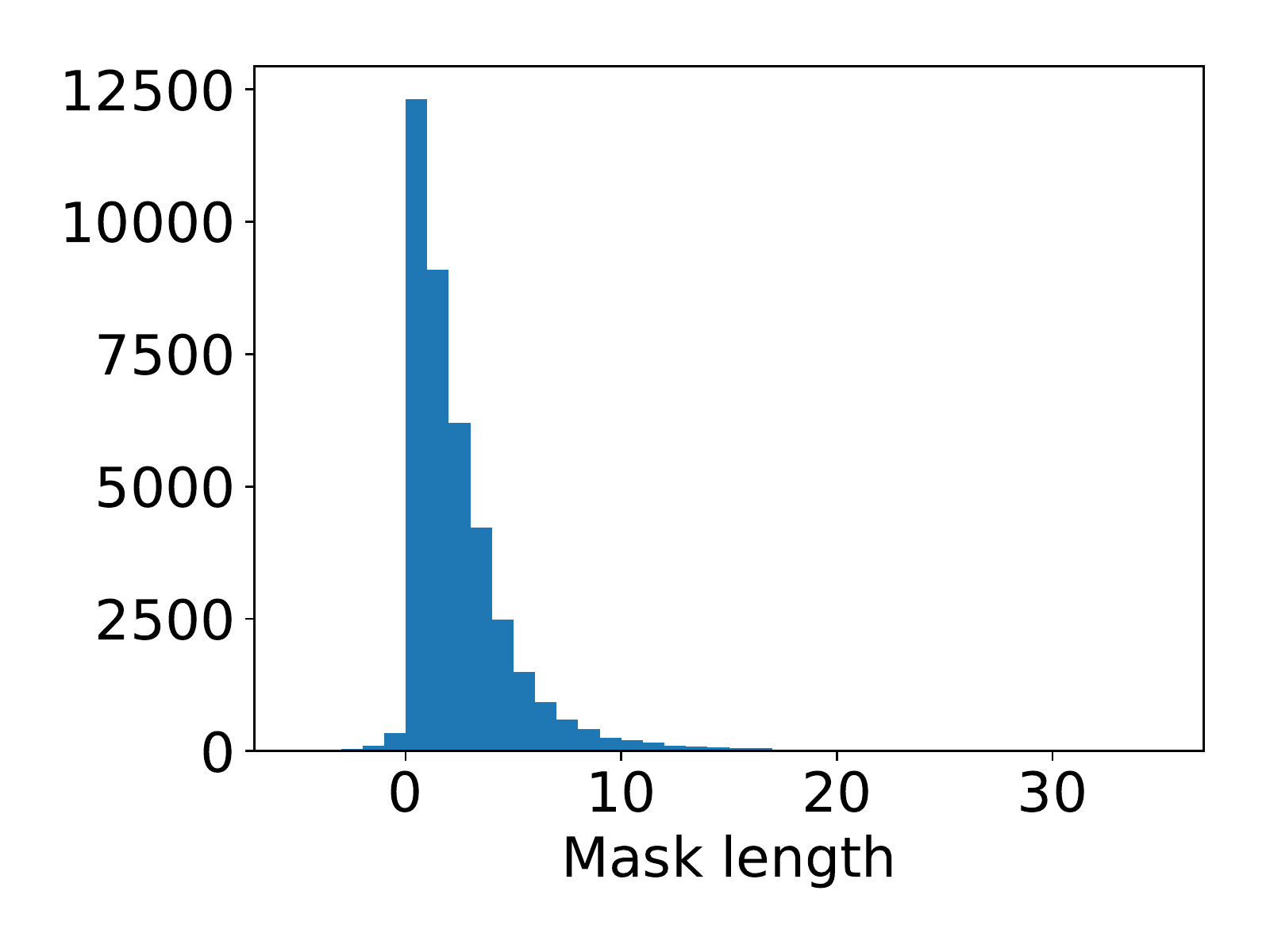}
  \caption{Masks for \emph{lm-greedy} strategy, $n=1, k=1$.}
\end{subfigure}
\begin{subfigure}{0.5\textwidth}
  \includegraphics[width=\linewidth]{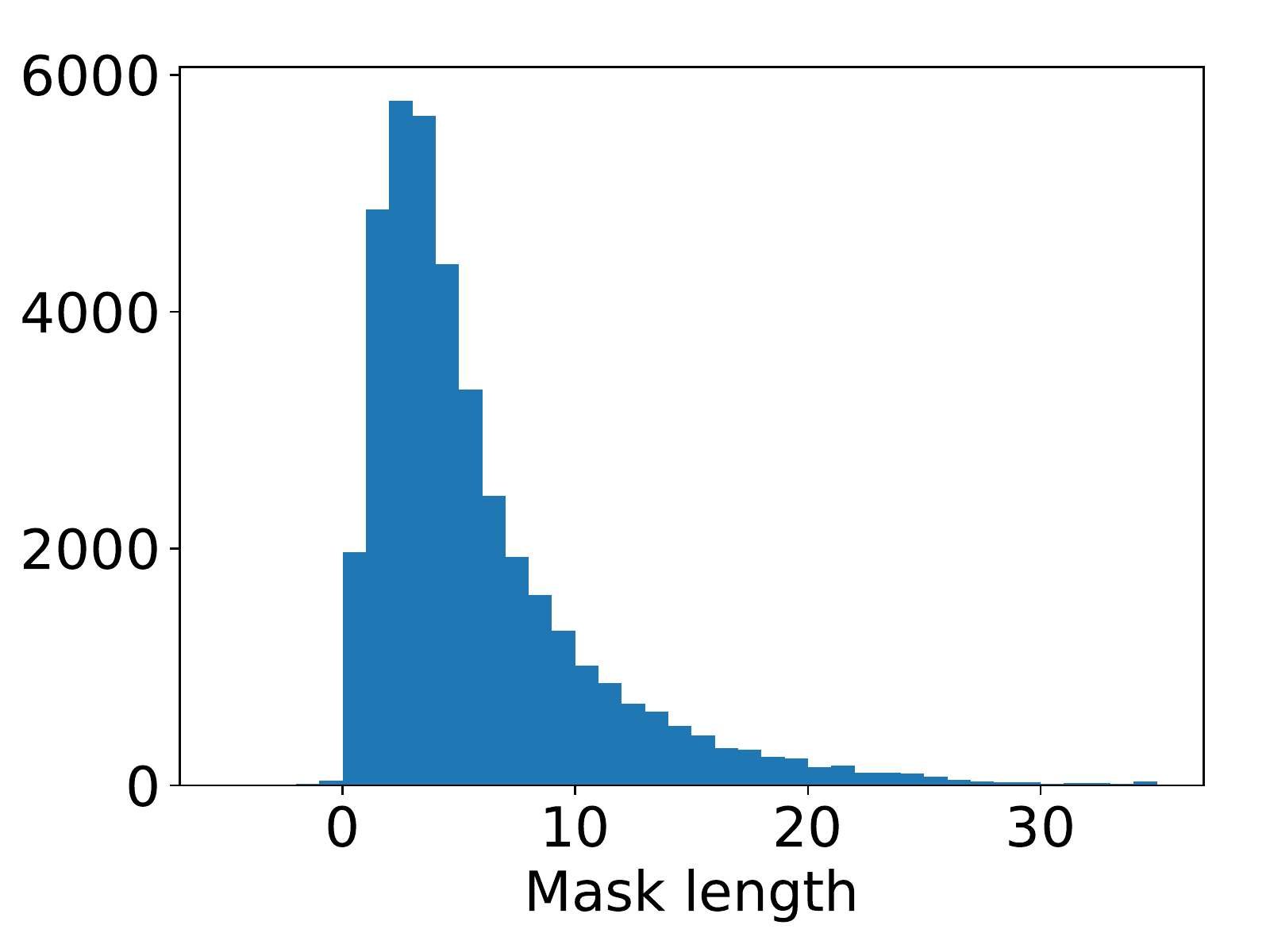}
  \caption{Masks for \emph{random} strategy, $n=3, k=5$.}
\end{subfigure}
\caption{Histograms of mask lengths over all prefixes for $en-de$ test set. The strategy on the left has low latency
but higher erasure (AL=1.54, NE=3.83) whereas the one on the right has low erasure but higher latency (AL=11.2, NE=0.22). 
In both cases we see that the vast majority of prefixes have low masks; in other words the skewed distribution means
most are below the mean.}
\label{fig:masks}
\end{figure}

We note that, whilst Figure~\ref{fig:dyn_mask} shows that our dynamic mask approach Pareto-dominates the mask-$k$,
it does not tell the
full story since it averages both the AL and NE across the whole corpus. To give more detail on the distribution of 
predicted masks, we show histograms of all masks on all prefixes for two of the source  prediction strategies
in Figure~\ref{fig:masks}. We select the source extension strategies at opposite ends of the curve, i.e. the highest AL / lowest NE strategy
and the lowest AL / highest NE strategy, and show results from en$\rightarrow$de.
We can see from the graphs in Figure~\ref{fig:masks} that the distribution across masks is highly skewed, with majority of prefixes having 
low masks, below the mean. This is in contrast with the mask-$k$ strategy where the mask has a fixed value, except at 
the end of a sentence where there is no mask. Our strategy is able to predict, with a reasonable degree of success,
when it is safe to reduce the mask.

\begin{figure}[ht!]
    \includegraphics[width=\linewidth]{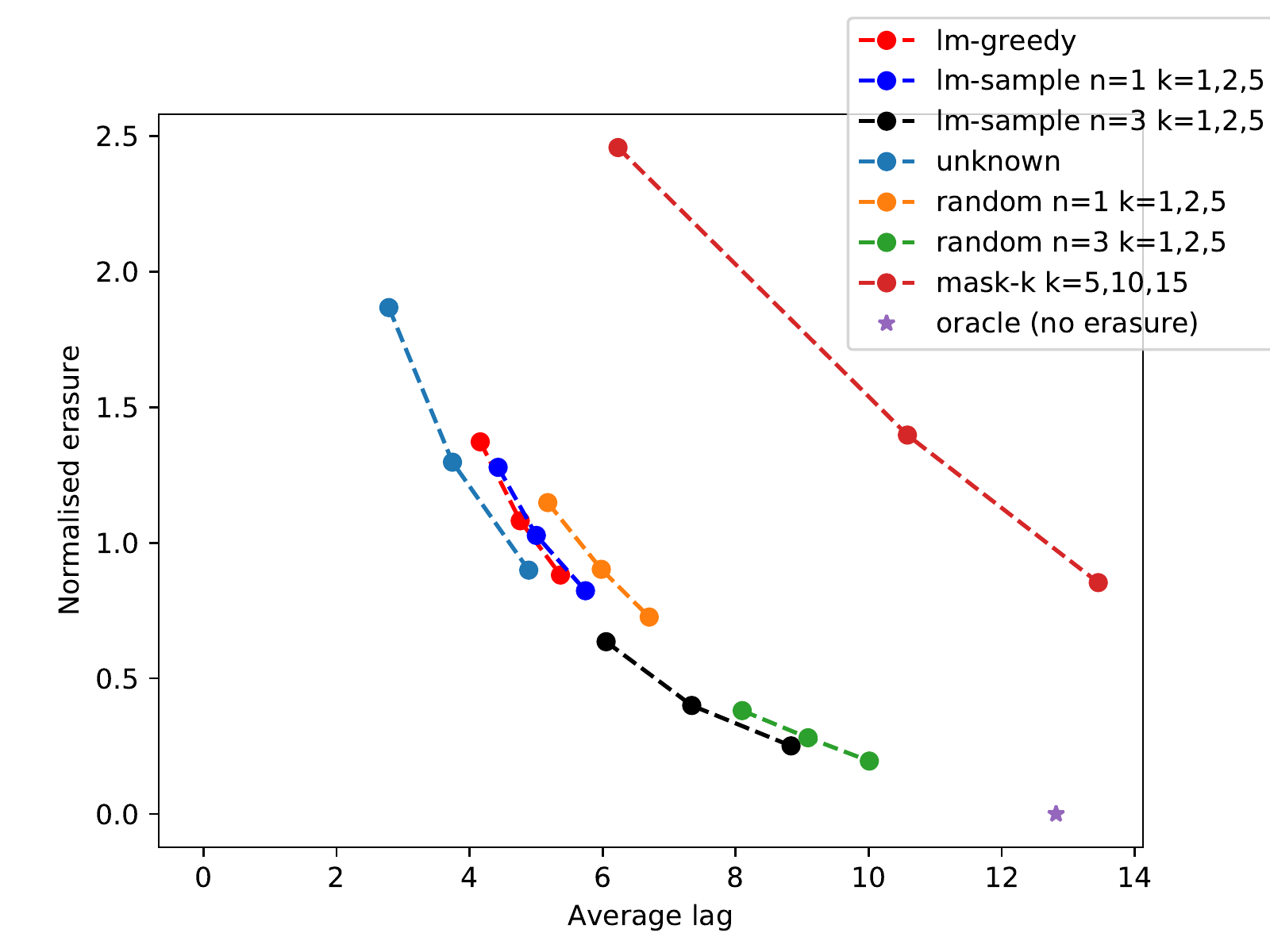}
    \caption{Effect of dynamic mask with different strategies, applied to a larger-scale en$\rightarrow$zh system.}
    \label{fig:dyn_mask_large}
\end{figure}

In order to provide further experimental verification of our retranslation model, we apply
the strategy to a larger-scale English$\rightarrow$Chinese system.
Specifically, we use a model that was trained on the entire parallel training data for the WMT20 en-zh
task\footnote{\url{www.statmt.org/wmt20/translation-task.html}}, in addition to the TED 
corpus used above. For the larger model we prepared as before, except with 30k BPE merges separately on each
language, and then we train a transformer-regular using Marian. By using this, we show that our strategy is able
to be easily slotted in a different model from a different translation toolkit.
The results are 
shown in Figure \ref{fig:dyn_mask_large}. We can verify that the pattern is unchanged in this setting.

\subsubsection{Dynamic Masking and Biased Beam Search}

\begin{figure}
\begin{subfigure}{0.5\textwidth}
  \includegraphics[width=\linewidth]{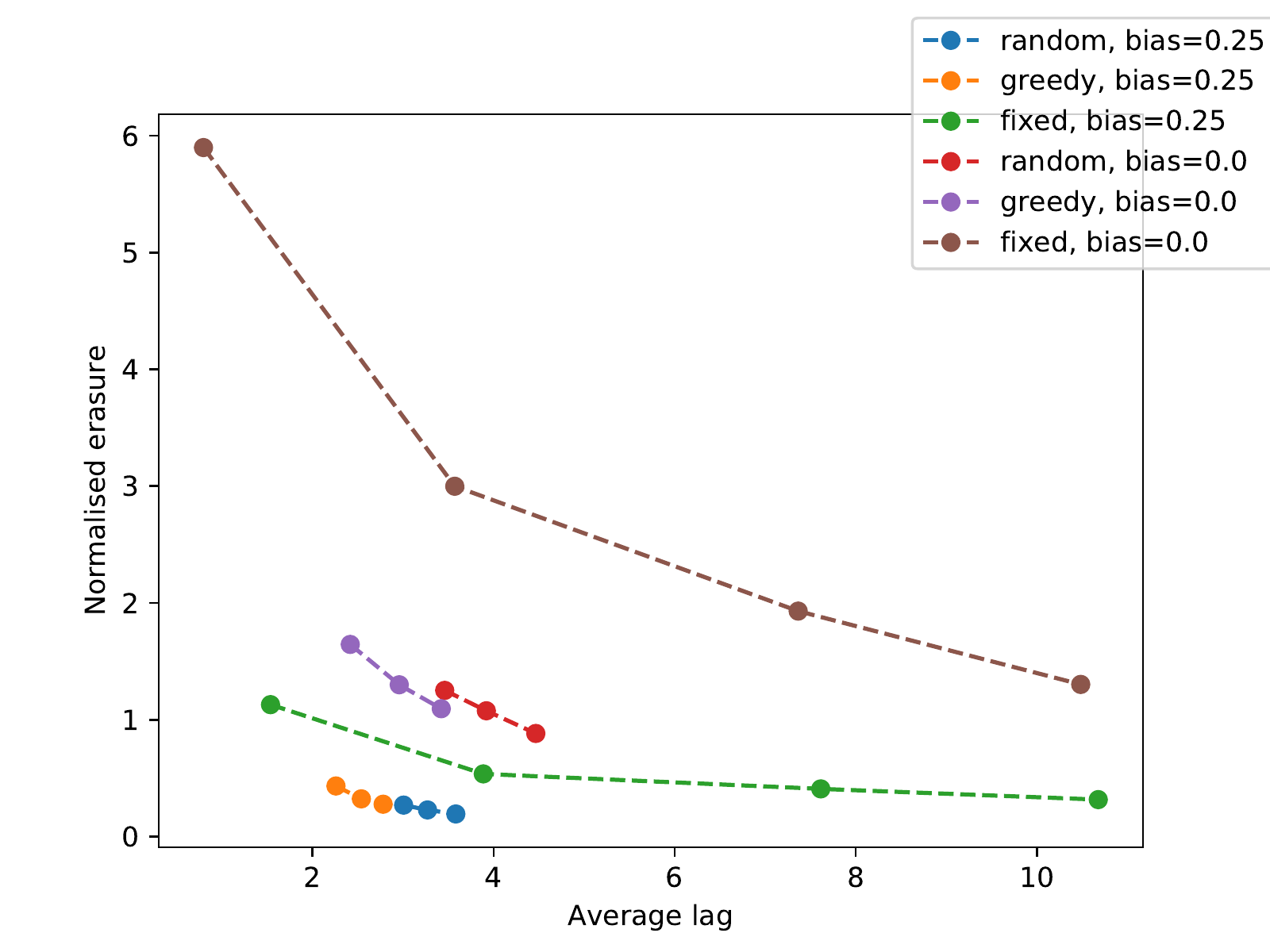}
  \caption{}
\end{subfigure}
\begin{subfigure}{0.5\textwidth}
  \includegraphics[width=\linewidth]{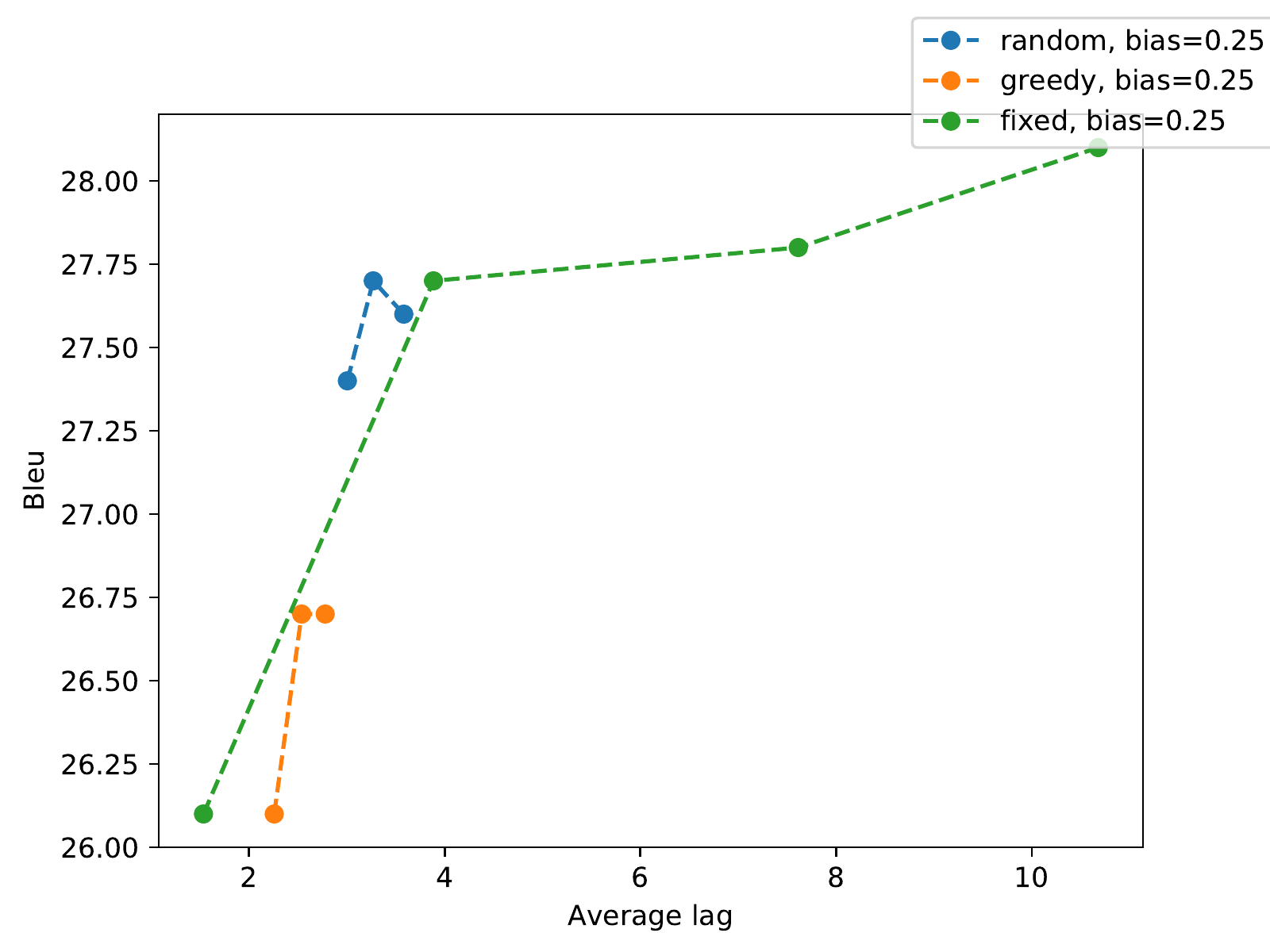}
  \caption{}
\end{subfigure}
\caption{Comparison of fixed mask with dynamic mask (random and greedy) with and without biased beam search. The figure
on the left shows the latency-flicker trade-off, whilst on the right we show the effect on \textsc{bleu} of adding the bias. With
bias=$0.0$, the \textsc{bleu} is 28.1.}
\label{fig:dynbias}
\end{figure}

We also note that dynamic masking can easily be combined with biased beam search. To do this, we perform all translation
using the model in Equation \ref{eqn:bias}, biasing translation towards the output (after masking) of the translation of the 
previous prefix. This has the disadvantage of requiring modification to the MT inference code, but 
figure \ref{fig:dynbias}(a) shows that combining biased beam search and dynamic
mask improves the flicker-latency trade-off curve, compared to using either method on its own.
In Figure \ref{fig:dynbias}(b) we show the effect on \textsc{bleu} of adding the bias. We can see that, using the the \emph{lm-greedy}
strategy, there is a drop in \textsc{bleu} (about -1), but that using the \emph{random} strategy we achieve a similar \textsc{bleu} as with the
fixed mask.

\subsubsection{Examples}

\begin{figure}[!ht]
\centering
\begin{tabular}{r|ll|l}
& Source & Translation & Output \\
\hline
prefix & Here & Hier sehen sie es &\\
extension & Here's & Hier ist es & Hier  \\
\hline
prefix & Here are & Hier sind sie &\\
extension & Here are some & Hier sind einige davon & Hier sind \\
\hline
prefix & Here are two & Hier sind zwei davon &\\
extension & Here are two things & Hier sind zwei Dinge & Hier sind zwei \\
\hline
prefix & Here are two patients & Hier sind zwei Patienten  &\\
extension & Here are two patients.    & Hier sind zwei Patienten.  & Hier sind zwei Patienten     \\
\end{tabular}
\caption{In this example, the dynamic mask covers up the fact that the MT system is unstable on short prefixes. Using the translation of the LM's predicted extension of the source, the system can see that the translation is uncertain, and so applies
a mask.
}
\label{tab:ex3}
\end{figure}

\begin{figure}[!ht]
\centering
\begin{tabular}{r|ll|l}
& Source & Translation & Output \\
\hline
prefix & But you know & Aber Sie wissen es  &\\
extension & But you know, & Aber wissen Sie, sie wissen schon & Aber \\
\hline
prefix & But you know what? & Aber wissen Sie was? & Aber wissen Sie was? \\
\end{tabular}
\caption{Here the dynamic masking is able to address the initial uncertainty between a question and statement,
which can have different word orders in German. The translation of the initial prefix and its extension produce different
word orders, so the system conservatively outputs ``Aber'', until it receivers further input.}
\label{tab:ex1}
\end{figure}

\begin{figure}[!ht]
\centering
{\footnotesize
\begin{tabular}{r|p{3.5cm}p{3.5cm}|p{3.5cm}}
& Source & Translation & Output \\
\hline
prefix & to paraphrase : it's not the strongest of  &  Um zu Paraphrasen: Es ist nicht die St\"arke der Dinge   & \\
extension & to paraphrase : it's not the strongest of the   & Um zu paraphrasen: Es ist nicht die St\"arke der Dinge  &Um zu paraphrasen: Es ist  nicht die St\"arke der Dinge  \\
\hline
prefix & to paraphrase : it's not the strongest of the  &  \textcolor{red}{Um zu} Paraphrasen: Es ist nicht die St\"arke der Dinge   & \\
extension & to paraphrase : it's not the strongest of the world   & \textcolor{red}{Zum}  paraphrasen: Es ist nicht die  st\"arksten der Welt  &Um zu paraphrasen: Es ist nicht die St\"arke der Dinge  \\
\end{tabular}
}
\caption{In this example we illustrate the refinement to the dynamic masking method, that is used typically when there are
small uncertainties early in the sentence. For the second prefix, its translation and the translation of its
extension differ at the beginning (``um zu'' vs ``zum''). Applying the LCP rule would result in a large mask, but instead
we output the translation of the previous prefix}
\label{tab:ex2}
\end{figure}

We illustrate the improved stability preovided by our masking strategy, by drawing three examples from the
 English$\rightarrow$German system's translation of our test set, in Figures \ref{tab:ex3}--\ref{tab:ex2}. In each example we show the source prefix (as provided by 
the ASR), its translation, its extension (in these examples, provided by the \emph{lm-greedy} strategy with $n=1, k=1$),
the translation of its extension, and the output after applying the dynamic mask.

In the first example (Figure~\ref{tab:ex3}) the MT system produces quite unstable translations of short prefixes, but the dynamic masking
algorithm detects this instability and applies short masks to cover it up. Since the translation of the prefix
and the translation of its extension differ, a short mask is applied and the resulting translation avoids erasure.

In the second example in Figure~\ref{tab:ex1}, there is genuine uncertainty in the translation because the fragment
 ``But you know'' can be translated
both as subject-verb and verb-subject in German, depending on whether its continuation is a statement or a question. The extension
predicted by the LM is just a comma, but it is enough to reveal the uncertainty in the translation and mean that the
algorithm applies a mask.

The last example (Figure~\ref{tab:ex2}) shows the effect of our refinement to the dynamic masking method, in reducing spurious masking. For
the second prefix, the 
translations of the prefix and its extension differ at the start, between ``Um zu'' and ``Zum''. Instead of masking to the
LCP (which would mask the whole sentence in this case) the dynamic mask algorithm just sticks with the output from the previous prefix.

%
%
%
%
\section{Conclusion}
We propose a dynamic mask strategy to improve the stability 
for the retranslation method in online SLT. We have shown that combining biased beam search with mask-$k$ works well in retranslation systems, but biased beam search requires a modified inference algorithm and
 hurts the quality and additional mask-$k$ used to reduce this effect gives a high latency. Instead, the dynamic mask strategy maintains the translation quality but gives a much better latency--flicker trade-off curve than mask-$k$.
 Our experiments also show that the effect of this strategy depends on both the length and number of predicted extensions, but the quality of predicted extensions is less important. Further experiments show that our method can be combined with biased beam search to give an improved latency-fliker trade-off, with a small reduction in \textsc{bleu}.

\vskip 1cm

\section*{Acknowledgements}
\lettrine[image=true, lines=2, findent=1ex, nindent=0ex, loversize=.15]{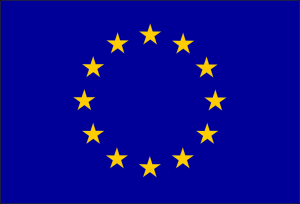}%
This work has received funding from the European Union’s Horizon 2020 Research and
Innovation Programme under Grant Agreement No 825460 (Elitr).

\bibliography{refs}

\begin{thebibliography}{}

\bibitem[Arivazhagan et~al., 2020a]{arivazhagan2019retranslation}
Arivazhagan, N., {Cherry}, C., {I}, T., {Macherey}, W., {Baljekar}, P., and
  {Foster}, G. (2020a).
\newblock {Re-Translation Strategies For Long Form, Simultaneous, Spoken
  Language Translation}.
\newblock In {\em ICASSP 2020 - 2020 IEEE International Conference on
  Acoustics, Speech and Signal Processing (ICASSP)}.

\bibitem[Arivazhagan et~al., 2019]{arivazhagan-etal-2019-monotonic}
Arivazhagan, N., Cherry, C., Macherey, W., Chiu, C.-C., Yavuz, S., Pang, R.,
  Li, W., and Raffel, C. (2019).
\newblock {Monotonic Infinite Lookback Attention for Simultaneous Machine
  Translation}.
\newblock In {\em Proceedings of the 57th Annual Meeting of the Association for
  Computational Linguistics}, pages 1313--1323, Florence, Italy. Association
  for Computational Linguistics.

\bibitem[Arivazhagan et~al., 2020b]{arivazhagan2020retranslation}
Arivazhagan, N., {Cherry}, C., {Macherey}, W., and {Foster}, G. (2020b).
\newblock {Re-translation versus Streaming for Simultaneous Translation}.
\newblock {\em arXiv e-prints}, page arXiv:2004.03643.

\bibitem[Bangalore et~al., 2012]{bangalore-etal-2012-real}
Bangalore, S., Rangarajan~Sridhar, V.~K., Kolan, P., Golipour, L., and Jimenez,
  A. (2012).
\newblock Real-time incremental speech-to-speech translation of dialogs.
\newblock In {\em Proceedings of the 2012 Conference of the North {A}merican
  Chapter of the Association for Computational Linguistics: Human Language
  Technologies}, pages 437--445, Montr{\'e}al, Canada. Association for
  Computational Linguistics.

\bibitem[Cettolo et~al., 2017]{Cettolo2017OverviewOT}
Cettolo, M., Federico, M., Bentivogli, L., Niehues, J., St{\"u}ker, S., Sudoh,
  K., Yoshino, K., and Federmann, C. (2017).
\newblock {Overview of the IWSLT 2017 Evaluation Campaign}.
\newblock In {\em Proceedings of IWSLT}.

\bibitem[Cho and Esipova, 2016]{DBLP:journals/corr/ChoE16}
Cho, K. and Esipova, M. (2016).
\newblock Can neural machine translation do simultaneous translation?
\newblock {\em CoRR}, abs/1606.02012.

\bibitem[Gu et~al., 2017]{gu-etal-2017-learning}
Gu, J., Neubig, G., Cho, K., and Li, V.~O. (2017).
\newblock Learning to translate in real-time with neural machine translation.
\newblock In {\em Proceedings of the 15th Conference of the {E}uropean Chapter
  of the Association for Computational Linguistics: Volume 1, Long Papers},
  pages 1053--1062, Valencia, Spain. Association for Computational Linguistics.

\bibitem[Junczys-Dowmunt et~al., 2018]{mariannmt}
Junczys-Dowmunt, M., Grundkiewicz, R., Dwojak, T., Hoang, H., Heafield, K.,
  Neckermann, T., Seide, F., Germann, U., Fikri~Aji, A., Bogoychev, N.,
  Martins, A. F.~T., and Birch, A. (2018).
\newblock Marian: Fast neural machine translation in {C++}.
\newblock In {\em Proceedings of ACL 2018, System Demonstrations}, pages
  116--121, Melbourne, Australia. Association for Computational Linguistics.

\bibitem[Koehn et~al., 2007]{koehn-EtAl:2007:PosterDemo}
Koehn, P., Hoang, H., Birch, A., Callison-Burch, C., Federico, M., Bertoldi,
  N., Cowan, B., Shen, W., Moran, C., Zens, R., Dyer, C., Bojar, O.,
  Constantin, A., and Herbst, E. (2007).
\newblock Moses: Open source toolkit for statistical machine translation.
\newblock In {\em Proceedings of the 45th Annual Meeting of the Association for
  Computational Linguistics Companion Volume Proceedings of the Demo and Poster
  Sessions}, pages 177--180, Prague, Czech Republic. Association for
  Computational Linguistics.

\bibitem[Ma et~al., 2019]{ma-etal-2019-stacl}
Ma, M., Huang, L., Xiong, H., Zheng, R., Liu, K., Zheng, B., Zhang, C., He, Z.,
  Liu, H., Li, X., Wu, H., and Wang, H. (2019).
\newblock {STACL}: Simultaneous translation with implicit anticipation and
  controllable latency using prefix-to-prefix framework.
\newblock In {\em Proceedings of the 57th Annual Meeting of the Association for
  Computational Linguistics}, pages 3025--3036, Florence, Italy. Association
  for Computational Linguistics.

\bibitem[Niehues et~al., 2018]{niehues2018inter}
Niehues, J., Pham, N.-Q., Ha, T.-L., Sperber, M., and Waibel, A. (2018).
\newblock Low-latency neural speech translation.
\newblock In {\em Proceedings of Interspeech}.

\bibitem[Post, 2018]{post-2018-call}
Post, M. (2018).
\newblock A call for clarity in reporting {BLEU} scores.
\newblock In {\em Proceedings of the Third Conference on Machine Translation:
  Research Papers}, pages 186--191, Brussels, Belgium. Association for
  Computational Linguistics.

\bibitem[Rangarajan~Sridhar et~al.,
  2013]{rangarajan-sridhar-etal-2013-segmentation}
Rangarajan~Sridhar, V.~K., Chen, J., Bangalore, S., Ljolje, A., and
  Chengalvarayan, R. (2013).
\newblock Segmentation strategies for streaming speech translation.
\newblock In {\em Proceedings of the 2013 Conference of the North {A}merican
  Chapter of the Association for Computational Linguistics: Human Language
  Technologies}, pages 230--238, Atlanta, Georgia. Association for
  Computational Linguistics.

\bibitem[Sennrich et~al., 2017]{sennrich-etal-2017-nematus}
Sennrich, R., Firat, O., Cho, K., Birch, A., Haddow, B., Hitschler, J.,
  Junczys-Dowmunt, M., L{\"a}ubli, S., Miceli~Barone, A.~V., Mokry, J., and
  N{\u{a}}dejde, M. (2017).
\newblock {N}ematus: a toolkit for neural machine translation.
\newblock In {\em Proceedings of the Software Demonstrations of the 15th
  Conference of the {E}uropean Chapter of the Association for Computational
  Linguistics}, pages 65--68, Valencia, Spain. Association for Computational
  Linguistics.

\bibitem[Sennrich et~al., 2016]{sennrich-haddow-birch:2016:P16-12}
Sennrich, R., Haddow, B., and Birch, A. (2016).
\newblock Neural machine translation of rare words with subword units.
\newblock In {\em Proceedings of the 54th Annual Meeting of the Association for
  Computational Linguistics (Volume 1: Long Papers)}, pages 1715--1725, Berlin,
  Germany. Association for Computational Linguistics.

\bibitem[Vaswani et~al., 2017]{vaswani2017attention}
Vaswani, A., Shazeer, N., Parmar, N., Uszkoreit, J., Jones, L., Gomez, A.~N.,
  Kaiser, {\L}., and Polosukhin, I. (2017).
\newblock Attention is all you need.
\newblock In {\em Advances in neural information processing systems}, pages
  5998--6008.

\bibitem[Zheng et~al., 2019a]{zheng-etal-2019-simpler}
Zheng, B., Zheng, R., Ma, M., and Huang, L. (2019a).
\newblock Simpler and faster learning of adaptive policies for simultaneous
  translation.
\newblock In {\em Proceedings of the 2019 Conference on Empirical Methods in
  Natural Language Processing and the 9th International Joint Conference on
  Natural Language Processing (EMNLP-IJCNLP)}, pages 1349--1354, Hong Kong,
  China. Association for Computational Linguistics.

\bibitem[Zheng et~al., 2019b]{zheng-etal-2019-simultaneous}
Zheng, B., Zheng, R., Ma, M., and Huang, L. (2019b).
\newblock Simultaneous translation with flexible policy via restricted
  imitation learning.
\newblock In {\em Proceedings of the 57th Annual Meeting of the Association for
  Computational Linguistics}, pages 5816--5822, Florence, Italy. Association
  for Computational Linguistics.

\end{thebibliography}
\bibliographystyle{apalike}

\end{document}